\documentclass[sigconf]{acmart}
\AtBeginDocument{%
  \providecommand\BibTeX{{%
    \normalfont B\kern-0.5em{\scshape i\kern-0.25em b}\kern-0.8em\TeX}}}

\copyrightyear{2024}
\acmYear{2024}
\setcopyright{acmlicensed}
\acmConference[WWW '24] {Proceedings of the ACM Web Conference 2024}{May 13--17, 2024}{Singapore, Singapore.}
\acmBooktitle{Proceedings of the ACM Web Conference 2024 (WWW '24), May 13--17, 2024, Singapore, Singapore}
\acmISBN{979-8-4007-0171-9/24/05}
\acmDOI{10.1145/3589334.3645670}

\usepackage{soul}
\usepackage{amsmath}
\usepackage{amsfonts}
\usepackage{amsthm}
\usepackage{booktabs}
\usepackage{xspace}
\usepackage{subfigure}
\usepackage{pifont}
\usepackage{mathtools}
\usepackage{environ}
\usepackage{multirow}
\usepackage{pdfpages}
\usepackage{balance}

\usepackage{algorithm}
\usepackage{algorithmic}
\newcommand{\model}{LLMQA\xspace}

\newcommand{\eg}{\emph{e.g.,}\xspace}

\DeclareMathOperator*{\argmax}{argmax}

\begin{document}

%%
%% The "title" command has an optional parameter,
%% allowing the author to define a "short title" to be used in page headers.
\title{Harnessing Multi-Role Capabilities of Large Language Models for Open-Domain Question Answering}

%%
%% The "author" command and its associated commands are used to define
%% the authors and their affiliations.
%% Of note is the shared affiliation of the first two authors, and the
%% "authornote" and "authornotemark" commands
%% used to denote shared contribution to the research.

\author{Hongda Sun}\authornote{Equal contribution.}
% \orcid{0000-0003-4850-6134}
\affiliation{
  % \department{Gaoling School of Artificial Intelligence}
  % \department{Gaoling School of AI (GSAI)}
  % \institution{Renmin University of China}
  \institution{Gaoling School of Artificial Intelligence, Renmin University of China}
  \city{Beijing}
  \country{China}}
\email{sunhongda98@ruc.edu.cn}

\author{Yuxuan Liu}\authornotemark[1]
\affiliation{
  \institution{Nankai University}
  \city{Tianjin}
  \country{China}
}
\email{2012677@mail.nankai.edu.cn}

\author{Chengwei Wu}
\affiliation{
  \institution{Beijing Academy of Artificial Intelligence}
  \city{Beijing}
  \country{China}
}
\email{wuchengwei0122@gmail.com}

\author{Haiyu Yan}
\affiliation{
  \institution{Renmin University of China}
  \city{Beijing}
  \country{China}}
\email{yanhaiyu@ruc.edu.cn}

\author{Cheng Tai}
\affiliation{
  \institution{Moqi, Inc.}
  \city{20 Emerald Hill Road}
  \country{Singapore}}
\email{chengt@myscale.com}

\author{Xin Gao}\authornotemark[2]
\affiliation{
  \institution{King Abdullah University of Science and Technology}
  \city{Thuwal}
  \country{Saudi Arabia}
}
\email{xin.gao@kaust.edu.sa}

\author{Shuo Shang}
\affiliation{
  \institution{University of Electronic Science and Technology of China}
  \city{Chengdu}
  \country{China}
}
\email{jedi.shang@gmail.com}

\author{Rui Yan}\authornote{Corresponding authors: Xin Gao and Rui Yan.}
\affiliation{
  % \department{Gaoling School of Artificial Intelligence}
  % \department{Gaoling School of AI (GSAI)}
  % \institution{Renmin University of China}
  \institution{Gaoling School of Artificial Intelligence, Renmin University of China}
  \city{Beijing}
  \country{China}}
\email{ruiyan@ruc.edu.cn}

%%
%% By default, the full list of authors will be used in the page
%% headers. Often, this list is too long, and will overlap
%% other information printed in the page headers. This command allows
%% the author to define a more concise list
%% of authors' names for this purpose.
\renewcommand{\shortauthors}{Hongda Sun et al.}

%%
%% The abstract is a short summary of the work to be presented in the
%% article.
\begin{abstract}
Open-domain question answering (ODQA) has emerged as a pivotal research spotlight in information systems.
Existing methods follow two main paradigms to collect evidence: (1) The \textit{retrieve-then-read} paradigm retrieves pertinent documents from an external corpus; and (2) the \textit{generate-then-read} paradigm employs large language models (LLMs) to generate relevant documents. 
However, neither can fully address multifaceted requirements for
evidence.
To this end, we propose \model, a generalized framework that formulates the ODQA process into three basic steps: query expansion, document selection, and answer generation, 
combining the superiority of both retrieval-based and generation-based evidence.
Since LLMs exhibit their excellent capabilities to accomplish various tasks,
we instruct LLMs to play multiple roles as generators, rerankers, and evaluators within our framework, integrating them to collaborate in the ODQA process.
Furthermore, we introduce a novel prompt optimization algorithm to refine role-playing prompts and steer LLMs to produce higher-quality evidence and answers. 
Extensive experimental results on widely used benchmarks (NQ, WebQ, and TriviaQA) demonstrate that \model achieves the best performance in terms of both answer accuracy and evidence quality, showcasing its potential for advancing ODQA research 
and applications.
\end{abstract}

%%
%% The code below is generated by the tool at http://dl.acm.org/ccs.cfm.
%% Please copy and paste the code instead of the example below.

% \begin{CCSXML}
% <ccs2012>
% <concept>
% <concept_id>10010147.10010178.10010179.10010182</concept_id>
% <concept_desc>Computing methodologies~Natural language generation</concept_desc>
% <concept_significance>500</concept_significance>
% </concept>
% <concept>
% <concept_id>10002951.10003227.10003351</concept_id>
% <concept_desc>Information systems~Data mining</concept_desc>
% <concept_significance>500</concept_significance>
% </concept>
% </ccs2012>
% \end{CCSXML}

% \ccsdesc[500]{Computing methodologies~Natural language generation}
% \ccsdesc[500]{Information systems~Data mining}



%%
%% Keywords. The author(s) should pick words that accurately describe
%% the work being presented. Separate the keywords with commas.
% \keywords{Do, Not, Us, This, Code, Put, the, Correct, Terms, for,
%   Your, Paper}

%% A "teaser" image appears between the author and affiliation
%% information and the body of the document, and typically spans the
%% page.
% \begin{teaserfigure}
%   \includegraphics[width=\textwidth]{sampleteaser}
%   \caption{Seattle Mariners at Spring Training, 2010.}
%   \Description{Enjoying the baseball game from the third-base
%   seats. Ichiro Suzuki preparing to bat.}
%   \label{fig:teaser}
% \end{teaserfigure}

% \received{20 February 2007}
% \received[revised]{12 March 2009}
% \received[accepted]{5 June 2009}

%%
%% This command processes the author and affiliation and title
%% information and builds the first part of the formatted document.
\maketitle

\section{Introduction}

In the interdisciplinary realm of information system applications, open-domain question answering (ODQA) has emerged as a pivotal research spotlight. This task is intrinsically knowledge-intensive and focuses on answering factoid questions, enhancing its utility to transcend the constraints of predefined domains~\cite{chen2017reading, petroni2020kilt, min2021neurips,gao2019product}.

Current ODQA methods follow two main paradigms in preparation for answering questions: (1) The \textit{retrieve-then-read} paradigm retrieves pertinent evidence documents from an external corpus and generates an answer based on them~\cite{karpukhin2020dense, izacard-grave-2021-leveraging}. Since retrieval models often rely on well-curated corpora like Wikipedia, they can provide highly factual and accurate information about the question; (2) The \textit{generate-then-read} paradigm directly employs language models to generate virtual documents~\cite{yu2023generate}, diversifying the evidence sources and enhancing answer coverage for the question.

Despite the individual merits of both paradigms, neither can adequately address the multifaceted requirements for evidence.
An intuitive solution is to integrate the strengths of these paradigms to collect evidence that combines factual reliability with diversity.
To this end, we propose \model, a novel generalized framework that incorporates the strengths of retrieval-based and generation-based evidence.
Specifically, we formulate the ODQA process into three fundamental steps: (1) \textbf{Query expansion} involves expanding the question by producing background passages or explanations, serving as generated-based evidence to enrich the context; (2) \textbf{Document selection} integrates retrieval-based evidence by reranking the retrieved documents, increasing their relevance to the answer; (3) \textbf{Answer generation} proceeds to generate the final answer based on comprehension of the question and obtained evidence.

To implement each step of ODQA, previous methods typically train specialized models for individual modules to obtain evidence documents and final answers~\cite{chuang2023expand}.
Limited by the inherent capabilities of these models, jointly optimizing each module to improve overall performance remains challenging.
Recent works have showcased the exceptional capabilities of large language models (LLMs) across various tasks~\cite{bubeck2023sparks}.
Specifically for ODQA, which requires integration of text generation~\cite{qin2023chatgpt, sun2023beamsearchqa, yu2023generate, chuang2023expand}, document ranking~\cite{ma2023large, chuang2023expand, ma2023zero}, and candidate evaluation~\cite{bohnet2022attributed, yue2023automatic}, the multiple aspects of capabilities of LLMs into each module.
Therefore, we aim to instruct LLMs to play three roles within our proposed unified framework: generators, rerankers, and evaluators.
By closely coordinating these roles and fostering collaboration among them, we can fully exploit their potential to enhance overall performance.
As shown in Figure~\ref{fig:roles}:
The \textbf{\textit{generator}} expands the query and provides comprehensive and pertinent information for answer generation; 
The \textbf{\textit{reranker}} prioritizes retrieved documents to distill more valid and relevant documents as evidence;
The \textbf{\textit{evaluator}} engages in interacting with the generator and reranker, providing evaluative feedback to refine their outputs.

The quality of LLMs in playing their distinct roles hinges on the quality of the prompts used to define tasks and guide their behaviors.
Therefore, the precision of obtained evidence is also sensitive to these prompts.
To better automatically design prompts, we present a novel prompt optimization algorithm to enhance the performance of LLMs across various roles within our framework.
During the ODQA generation process, we treat evidence (\eg query expansion and selected documents) as latent variables and leverage variational inference to optimize role-playing prompts, guiding LLMs toward producing higher-quality evidence and answers.

We conduct experiments on widely used ODQA benchmarks: NQ, WebQ, and TriviaQA. The results show that our \model advances the state-of-the-art performance on both answer accuracy and evidence quality. Compared with baselines, \model achieves remarkable improvement in EM scores (4.0@TriviaQA, 2.7@WebQ, 3.1@NQ), demonstrating the effectiveness of multi-role LLMs for ODQA. The role of query expansion generator can achieve 73\%,76\%, and 87\% recall scores for the answer in generated expansions. The role of reranker increases answer coverage by about 8.1\%.

To sum up, our main contributions can be summarized as follows:

$\bullet$ We propose \model, a generalized framework model to formulate the ODQA process, which is a novel paradigm to combine the strengths of retrieval-based and generation-based evidence.

$\bullet$ 
We effectively instruct LLMs to play three roles of generators, rerankers, and evaluators respectively, and integrate their collaborative interactions under our proposed unified framework,

$\bullet$ We introduce a novel prompt optimization algorithm to guide LLMs in producing higher-quality evidence and answers. Extensive experimental results show that \model advances the best performance in terms of both answer accuracy and evidence quality.

\section{Related Work}

\subsection{Open-Domain Question Answering}

For collecting the related documents as evidence, existing methods can be categorized into the following two main paradigms:

\vspace{0.5em}
\noindent \textbf{Retrieve-then-read paradigm.} Pioneered by \cite{chen2017reading}, most recent approaches consist of two main modules: 
The \textit{retriever} first retrieves documents relevant to the given question from an external knowledge base. The \textit{reader} then comprehends questions and retrieved documents and generates the corresponding answer. One branch focuses on improving the \textit{retriever}. Sparse retrieval with inverted indexes (\eg TF-IDF or BM25)  is generally used in traditional approaches~\cite{robertson2009probabilistic}. Dense retrieval using language models such as ORQA \cite{lee2019latent}, DPR \cite{karpukhin2020dense}, RocketQA \cite{qu2020rocketqa}, ColBertQA \cite{khattab2021relevance}, and ART \cite{sachan2023questions} becomes dominant. The other branch focuses on enhancing the comprehension ability of \textit{reader} to generate more accurate answers \cite{izacard-grave-2021-leveraging, cheng-etal-2021-unitedqa}. With the development of LLMs, most \textit{readers} are adopted from fine-tuned T5 \cite{raffel2020exploring} or InstructGPT \cite{ouyang2022training}.

\vspace{0.5em}
\noindent \textbf{Generate-then-read paradigm.} Previous works have demonstrated that the knowledge preserved in LLMs can serve as a ``\textit{generative retriever}'' \cite{radford2019language, petroni2019language, roberts-etal-2020-much}. Although many existing approaches adopt LLMs in ODQA, they cannot fully harness the generation capability of LLMs \cite{qin2023chatgpt, kamalloo2023evaluating, wang2023evaluating, sun2023beamsearchqa}. 
GenRead is the first to explore the potential of the generation-based evidence for ODQA, which instructs an LLM to generate documents based on clusters of question-document pairs and the given question~\cite{yu2023generate}. Then these generated documents and the question are fed into LLM together to produce the final answer.

Considering the limitations of a single paradigm, we propose to seamlessly integrate retrieval-based and generation-based evidence, effectively harnessing the capabilities of LLMs.

\subsection{Capabilities of LLMs}
Recent advancements in model scales~\cite{chowdhery2022palm,ouyang2022training} provide
LLMs with impressive capabilities in text generation, ranking, and evaluation.

\vspace{0.5em}
\noindent \textbf{Generation capability of LLMs.} Recent studies have highlighted the superior text generation capability of LLMs in few-shot and zero-shot scenarios~\cite{brown2020language, chowdhery2022palm, zhou2023comprehensive, bubeck2023sparks}.
The knowledge stored in LLMs could be retrieved during inference~\cite{petroni2019language, roberts-etal-2020-much}. Hence, some studies directly prompt LLMs to generate answers to the question in ODQA~\cite{qin2023chatgpt, kamalloo2023evaluating, wang2023evaluating, sun2023beamsearchqa}. 
Other approaches utilize the generation capability to expand the query or enrich the context~\cite{mao2020generation, yu2023generate, chuang2023expand,doc2query}.

\vspace{0.5em}
\noindent \textbf{Ranking capability of LLMs.} Previous works show that compared to few-shot information extraction, 
LLMs are better at reranking for hard examples. 
\citeauthor{ma2023large} propose a \textit{filter-then-rerank} paradigm, which utilizes LLMs to rerank the candidates filtered by smaller language models to generate the final response. 
\citeauthor{chuang2023expand} apply LLMs to rerank expanded queries for better results. 
\citeauthor{ma2023zero} replace pointwise reranking with listwise reranking to reorder the list of documents based on the relevance to the query.

\vspace{0.5em}
\noindent \textbf{Evaluation capability of LLMs.}
LLMs are chosen as evaluators due to their robust comprehension and reasoning capabilities~\cite{bubeck2023sparks, chiang2023vicuna,Gao2023ConfuciusIT,sun2023indeterminacy}. \citeauthor{weng2022large} leverage the self-verification capability of LLMs for better reasoning. \citeauthor{shinn2023reflexion} use self-reflective feedback as a semantic gradient providing a concrete direction to learn from prior mistakes. \citeauthor{madaan2023self} present an iterative self-refinement algorithm that alternates between feedback and refinement. 
Additionally, LLMs are also used to evaluate attribution between generated answers and references~\cite{bohnet2022attributed, yue2023automatic}. 

In this paper, we aim to effectively integrate multi-role capabilities of LLMs to enhance the overall performance on ODQA.

\subsection{Prompt Optimization}

Previous works have emphasized that subtle differences in prompts could lead to tremendous performance degradation in generated results~\cite{gao2020making, zhou2022large, liu2023pre}. Consequently, prompt optimization has attracted great attention in recent years, with two primary approaches: manual design~\cite{reynolds2021prompt} or automatic generation~\cite{shin2020autoprompt}.
Gradient-based prompt tuning can optimize prompts embedding in a continuous space \cite{liu2021gpt, liu2021p}. 
In contrast, discrete prompt optimization has been extensively studied including prompt scoring \cite{davison2019commonsense}, prompt generation \cite{gao2020making} and prompt paraphrasing \cite{yuan2021bartscore}. Recently, \citeauthor{zhou2022large} propose APE for automatic prompt optimization by iteratively selecting prompt candidates to maximize the potential score functions. DLN~\cite{sordoni2023deep} steps further by viewing LLMs as language layers and prompts as learnable parameters.

Inspired by these methods, we present a novel prompt optimization algorithm to refine essential prompts for query expansion, document reranking, and answer generation, enabling LLMs to produce better evidence and answers.

\section{Method}
\subsection{Task Formulation}
Previous methods collect evidence by retrieving or generating relevant background passages or explanations to facilitate accurate answer identification~\cite{chen2017reading,karpukhin2020dense,ouyang2022training}.
Expanding upon this concept, we formulate the generation process of ODQA as the following three fundamental steps: (1) \textbf{Query expansion}: We commence with the input question, designated as query $q$. To enrich the context and improve document selection and answer generation, we utilize knowledge stored in language models to generate additional background information, denoted as query expansion $e$; 
(2) \textbf{Document selection}: Leveraging both the query $q$ and its expansion $e$, we initially retrieve the top-$n$ documents that are relevant to answering the question as candidates. Subsequently, we compare these candidates to prioritize those documents most likely to contain the answer.
Based on this criterion, we rerank these $n$ candidates and retain top-$k$ documents, represented as $d$, which collectively constitute the evidence in conjunction with query expansion $(e,d)$;
(3) \textbf{Answer generation}: Based on the query $q$ and the derived evidence $(e,d)$, we proceed to generate the final answer $a$ in response to the question with a reader model. 

\begin{figure*}
    \centering
    \includegraphics[width=0.8\textwidth]{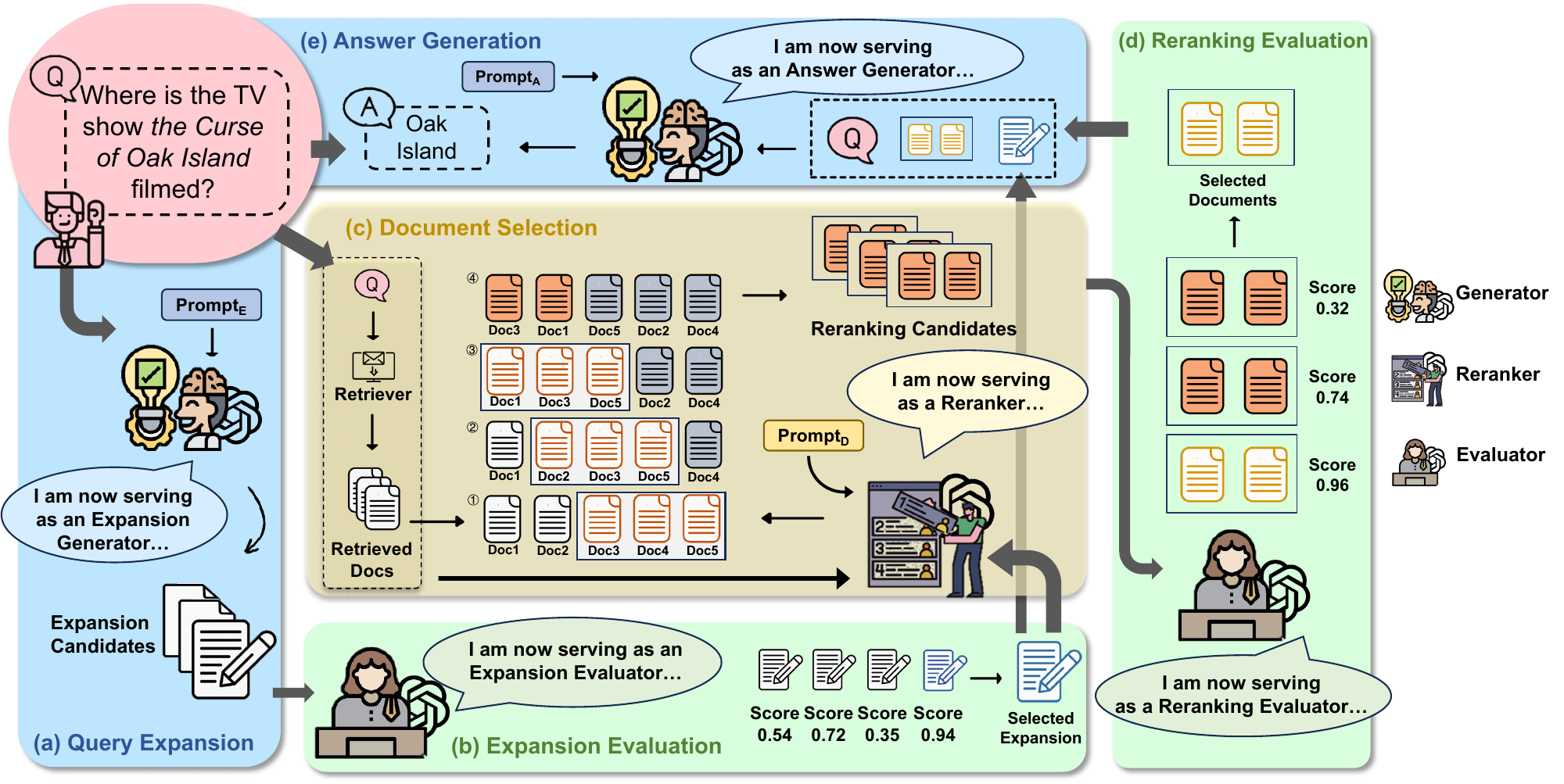}
    \setlength{\abovecaptionskip}{0.2cm}
    \setlength{\belowcaptionskip}{-0.2cm}
    \caption{The overview of our \model. Three different role-play LLMs execute five main steps: (a) generate query expansion according to the question by \textit{generator}; (b) select the best query expansion by \textit{evaluator}; (c) rerank the top-$k$ documents according to the question and generated expansion by \textit{reranker}; (d) select the best reranked documents by \textit{evaluator}; (e) generate answer according to the question, generated expansion and reranked documents by \textit{generator}. A more detailed insight into sliding window reranking: select top-2 documents from top-5 retrieved candidates with window size $w=3$, step $l=1$.}
    \label{fig:model}
\end{figure*}

Furthermore, this generation process can be effectively formulated using a Bayesian graphical model that aligns closely with the three aforementioned steps, parameterized by the following probability distribution:
\begin{equation}
    P(a|q) = \sum_{e}\sum_{d}P(e|q)P(d|q,e)P(a|q,e,d),\label{graph}
\end{equation}
where we consider the evidence $(e,d)$ as latent variables, which require to be optimized by maximizing this marginal likelihood.
Consequently, the acquisition of the most appropriate evidence for question answering becomes a critical aspect of this task.
Considering the prominent performance of LLMs on various tasks, we harness LLMs in multiple roles that collaborate with each other in the ODQA generation process.
The framework overview of \model is shown in Figure~\ref{fig:model}.
In the subsequent sections, we will introduce in detail how to leverage the multi-role capabilities of LLMs to enhance the ODQA task.

\begin{figure}
    \centering
    \includegraphics[width=0.8\linewidth]{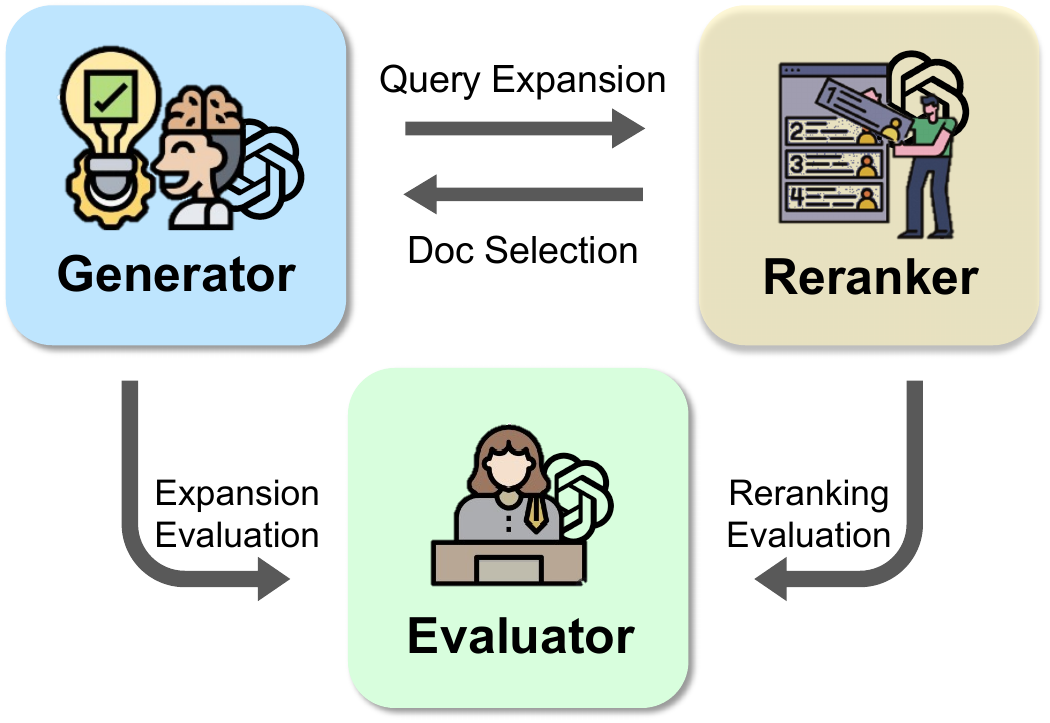}
    \setlength{\abovecaptionskip}{0.2cm}
    \setlength{\belowcaptionskip}{-0.3cm}
    \caption{Collaborative interactions of multiple LLM roles.}
    \label{fig:roles}
\end{figure}

\subsection{Query Expansion}
Generally, the questions posed in ODQA datasets are brief and concise, indicating that relying solely on the question itself as a query can lead to a substantial challenge: inadequate query context makes it difficult to support accurate document selection and answer generation. 
To address this challenge, 
we add a pivotal step known as \textit{query expansion} that aims to enrich the original question with a broader context.
The generated expansions are mainly used to analyze the key points required to answer a given question and provide sufficient background information for subsequent steps.
In this process, we instruct an LLM to play the role of generator leveraging its powerful context understanding and text generation capabilities.
Specifically, we employ an LLM-based expansion generator $G_e$ to facilitate the query expansion step.
Given a question $q$, its query expansion $e$ can be generated by 
\begin{equation}
    e=G_e(q;\theta_e),
\end{equation}
where $\theta_e$ represents the prompt to instruct the query expansion.

\subsection{Document Selection}

In addition to the query expansion, relevant documents are more commonly used as evidence to include accurate answers to the question.
To identify the most appropriate documents, we divide this document selection process into two distinct stages: 

\textbf{(1) Coarse-grained retrieval of top-$n$ documents:} 
we first retrieve a set of top-$n$ documents that are potentially relevant to the given question by employing established information retrieval techniques such as DPR~\cite{karpukhin2020dense} or BM25~\cite{chen2017reading}.
These retrieval methods provide an initial score for each candidate to describe the relevance between documents and questions.
However, such methods may not always capture nuanced semantic relationships between the query and documents, leading to false positives or irrelevant documents in the initial set.

\textbf{(2) Fine-grained reranking of top-$k$ documents from $n$ candidates:}
we proceed with the reranking of documents to ensure that those more likely to contain the answer are prioritized.
This stage involves comparing the documents to determine which ones exhibit higher quality and relevance to the query.
Inspired by LLM-based ranking approaches~\cite{ma2023large,chuang2023expand,ma2023zero}, We instruct LLM to play the role of document reranker $R_d$ for further screening out top-$k$ ($k<n$) documents from the initial pool of $n$ candidates. 
Considering the limitation on input tokens for LLMs, 
we iteratively rerank a subset of documents each time and complete the reranking of all candidates through a sliding window.
Specifically, we set the window size to $w$ and the step size to $l$.
We start from the last position of the initially sorted documents.
In each iteration, we focus on comparing $w$ documents within the sliding window and reorder the documents based on their likelihood of containing the answer.
With the sliding window moving forward
by $l$ steps, thus the top $w-l$ reranked documents in the original window are reserved and $l$ new documents are added, then the next $w$ documents can be reordered. 
This iterative process continues until the sliding window reaches the front, and we consider the first $k = w-l$ documents as the final evidence documents $d$. 
Overall, this document selection process can be simplified as:
\begin{equation}
    d = R_d(q,e;\theta_d),
\end{equation}
where $\theta_d$ denotes the prompt for $R_d$ to ensure that documents are ranked in alignment with the desired relevance and quality.

\subsection{Answer Generation}
Based on the query $q$, and the evidence $(e,d)$, the final step in ODQA is to generate the final answer with the integration and comprehension of pertinent information within the evidence.
The evidence can encompass essential information that directly provides the answer to the question, or it may comprise an analysis and explanation necessary for formulating the answer. Consequently, the central objective of answer generation is to employ a reader model for the systematic extraction and comprehension of valuable insights from the evidence context.
We utilize an LLM-based reader $G_a$ to generate a precise and dependable response as the predicted answer to the question, and formulate this process as:
\begin{equation}
    a = G_a(q,e,d;\theta_a),
\end{equation}
where $\theta_a$ indicates the prompt for answer generation to ensure that the generated answer can align with the context and requirements of the original question and its evidence.

\subsection{Evaluators for Generation and Reranking}
As shown in Figure~\ref{fig:roles}, evaluators also play a crucial role in query expansion and document reranking, engaging in a dynamic interaction with both the generator and reranker. Leveraging the advanced capabilities of LLMs to evaluate text quality under specific standards, we can instruct LLMs to play the role of evaluators to assess the performance of the generator and the reranker.
The primary objective of evaluators is to assign quality scores to multiple candidates generated by the generator and reranker. These scores reflect the likelihood that each candidate is appropriate and accurate for specific conditions or requirements.
For a given question, we employ the expansion evaluator $S_e$ to individually score each candidate expansion ranging from 0 to 1, which is used to assess the degree of their relevance and logical consistency.
Similarly, we use the reranking evaluator $S_r$ to score different top-$k$ reranking candidates 
generated by the reranker $R_d$, assessing the contribution of each ranking result to answering the question.
The scoring process of evaluators $S_e$ and $S_r$ can be formulated as:
\begin{align}
    s_{e_j} & = S_e(e_j; G_e(q)), \\
    s_{d_j} & = S_r(d_j; R_d(q,e)), 
\end{align}
where $s_{e_j}$ and $s_{d_j}$ represent the scores assigned to the $j$-th candidate of generated query expansion and reranked documents. These scores serve as critical metrics for evaluating the performance of the generator and reranker and further promoting overall generation and ranking capabilities of LLMs.

\begin{algorithm}[t]
\setlength{\abovecaptionskip}{0.1cm}
\setlength{\belowcaptionskip}{0cm}
\caption{Training Process for Prompt Optimization.}
\label{alg:algorithm}
\begin{flushleft}
\textbf{Input}: Training data: $X_{tra}$, LLM-based roles: $G_e$, $R_d$, $G_a$, $S_e$, $S_r$, $S_a$, backward updating functions: $U_e$, $U_d$, $U_a$. \\
\textbf{Parameters}: $\theta_e$, $\theta_d$, $\theta_a$. \\
\end{flushleft}
\begin{algorithmic}[1] %[1] enables line numbers
\FOR{$(q, a)$ in $X_{tra}$}
\STATE Generate the prior $\widetilde{e} = G_e(q;\theta_e)$ by Expansion Generator
\STATE Generate the prior $\widetilde{d} = R_d(q,\widetilde{e};\theta_d)$ by Document Reranker
\STATE Generate the prior $\widetilde{a} = G_a(q,\widetilde{e},\widetilde{d};\theta_a)$ by Answer Generator
\STATE Sample $n$ posterior $\widehat{d}_1, \widehat{d}_2,\cdots,\widehat{d}_n$ from $R_d(q,\widetilde{e},\widetilde{d},a;\phi_d)$
\STATE Score $s_{d_i} = S_r(\widehat{d}_i;R_d(q,\widetilde{e}))$, $s_{a_i} = S_a(a;G_a(q,\widetilde{e},\widehat{d}_i))$ for $\widehat{d}_i$ \\
\STATE Calculate posterior score $v_{d_i} = s_{d_i} * s_{a_i}$
\STATE Select the best posterior $\widehat{d}_*=\text{argmax}_{i}\{v_{d_i}\}$
\STATE Sample $m$ posterior $\widehat{e}_1, \widehat{e}_2,\cdots,\widehat{e}_m$ from $G_e(q,\widetilde{e},\widehat{d}_*,a;\phi_e)$
\STATE Score $s_{e_j} = S_e(\widehat{e}_j;G_e(q))$,
$s_{d_j} = S_r(\widehat{d}_*;R_d(q,\widehat{e}_j))$,
and $s_{a_j} = S_a(a;G_a(q,\widehat{e}_j,\widehat{d}_*))$ for each $\widehat{e}_j$
\STATE Calculate posterior score $v_{e_j} = s_{e_j}*s_{d_j}*s_{a_j}$
\STATE Select the best posterior $\widehat{e}_*=\text{argmax}_{j}\{v_{e_j}\}$
\STATE Sample $K$ candidates $\widehat{\theta}_{a_k}=U_a(q,\widetilde{e},\widetilde{d}, \widetilde{a}, a)$ for $\theta_a$
\STATE Select $\widehat{\theta}_{a_*} = 
\argmax\limits_{k}\sum\limits_{i=1}^{n}\sum\limits_{j=1}^m v_{e_j}v_{d_i}\log S_a(a;G_a(q,\widehat{e}_j,\widehat{d}_i;\widehat{\theta}_{a_k}))$
\STATE Sample $K$ candidates $\widehat{\theta}_{d_k}=U_d(q,\widetilde{e}, \widetilde{d}, \widehat{d}_*)$ for $\theta_d$
\STATE Select $\widehat{\theta}_{d_*} = \argmax\limits_{k}\sum\limits_{i=1}^{n}\sum\limits_{j=1}^m v_{e_j}v_{d_i}\log S_r(\widehat{d}_i;R_d(q,\widehat{e}_j;\widehat{\theta}_{d_k}))$
\STATE Sample $K$ candidates $\widehat{\theta}_{e_k}=U_e(q,\widetilde{e}, \widehat{e}_*)$ for $\theta_e$
\STATE Select $\widehat{\theta}_{e_*} = \argmax\limits_{k}\sum\limits_{j=1}^m v_{e_j}\log S_e(\widehat{e}_j;G_e(q;\widehat{\theta}_{e_k}))$
\STATE Update parameters: $\theta_a \leftarrow \widehat{\theta}_{a_*}$, \ $\theta_d \leftarrow \widehat{\theta}_{d_*}$, \ $\theta_e \leftarrow \widehat{\theta}_{e_*}$
\ENDFOR
\RETURN $\theta_e$, $\theta_d$, $\theta_a$. 
\end{algorithmic}
\end{algorithm}

\subsection{Prompt Optimization}
The role-play performance of the generator and reranker still heavily relies on the prompt design in each ODQA generation process.
Therefore, we explore how to design better role-play prompts or expansion generation $\theta_e$, document reranking $\theta_d$, and answer generation $\theta_a$ to fully exploit the potential of LLMs.
We propose a novel algorithm to enable prompt optimization under the unique graphical model structure of ODQA.
Throughout the ODQA generation process, we do not require the LLM parameters, but instead treat three natural language prompts as learnable parameters.
In Equation~\eqref{graph}, the distributions of latent variables $e$ and $d$ are determined by these prompts and need to be approximated by probabilistic inference techniques.
To ensure consistency with the graphical model, we propose to use variational inference to learn the hidden distributions and optimize prompts.
We denote the prior distribution as $P_\theta$ and the posterior distribution as $P_\phi$, and the original log-likelihood could be bounded by the following ELBO:
{\fontsize{6.95}{0}\selectfont
\begin{align}
    & \log P(a|q) \nonumber \\ 
    \ge & \sum_e\sum_d P_{\phi_e}(e|q,a)P_{\phi_d}(d|q,e,a)\log \frac{P_{\theta_e}(e|q)P_{\theta_d}(d|q,e)P_{\theta_a}(a|q,e,d)}{P_{\phi_e}(e|q,a)P_{\phi_d}(d|q,e,a)}.
\end{align}
}
As shown in Algorithm~\ref{alg:algorithm}, for the question $q$, 
we use predefined $G_e$, $R_d$ and $G_a$ to sequentially simulate the priors $P_{\theta_e}$, $P_{\theta_d}$, and $P_{\theta_a}$, and generate the query expansion $\widetilde{e}$, the reranked documents $\widetilde{d}$ and the predicted answer $\widetilde{a}$ during forward inference.
Next, to approximate the posteriors $P_{\phi_e}$ and $P_{\phi_d}$, we consider the following two aspects: (1) We add the ground-truth target as an additional condition to estimate the posteriors; (2) We sample several posterior candidates near the prior to ensure low Kullback–Leibler (KL) divergence between them in the space of discrete texts.
Denoting the prior reranked documents as $\widetilde{d} = (\widetilde{d}_1, \widetilde{d}_2,\cdots,\widetilde{d}_{k-1},\widetilde{d}_k)$, the $i$-th posterior reranked documents can be denoted as $\widehat{d}_i = (\widetilde{d}_1,\widetilde{d}_2,\cdots,\widetilde{d}_{k-1},\widetilde{d}_{k+i})$, where only the ``last document'' in the list is replaced.
Then we use evaluators $S_r$ and $S_e$ to score each posterior candidate for estimating $P_{\phi_d}$.
The best posterior $\widehat{d}_*$ among these candidates is selected as the current ``ground-truth'' reranking documents. 
The posterior query expansions are generated by a minor edit of the prior expansion, then the best posterior $\widehat{e}_*$ is selected by analogy.

Subsequently, we define a backward process to update prompts. For the answer generation prompt $\theta_a$, we sample $K$ candidates near it using an updating function $U_a(q,\widetilde{e}, \widetilde{d}, \widetilde{a}, a)$, to guide prompts to update in a direction that brings the predicted answer $\widetilde{a}$ closer to the actual answer $a$.
Then all the previous posterior candidates are used to estimate ELBO, and the best $\widehat{\theta}_{a_*}$ to maximize ELBO can be selected as the refined prompt for answer generation. 
Similar processes are introduced to refine prompts $\theta_d$ and $\theta_e$, while updating functions $U_d$ and $U_e$ are used to guide the directions to refine document reranking and query expansion.

\section{Experiments}

\begin{table*}
\setlength{\abovecaptionskip}{0.1cm}
\setlength{\belowcaptionskip}{-0.4cm}
\caption{Comparison results on TriviaQA, WebQ, and NQ datasets. Our EM scores are given by the mean of 10 rounds of bootstrapping sampling, with bold numbers indicating $p$-values below 0.01 under a significance test.}
\renewcommand\arraystretch{1}
\centering
\begin{tabular}{lccccc}
\hline
\multicolumn{1}{l|}{\textbf{Method}} &
  \textbf{\begin{tabular}[c]{@{}c@{}}\#Reader\\ parameters\end{tabular}} &
  \textbf{\#Documents} &
  \textbf{TriviaQA} &
  \textbf{WebQ} &
  \textbf{NQ} \\ \hline
\multicolumn{6}{l}{\textit{baselines without LLMs; $\dag$ was reported by paper, $*$ was reproduced by ours.}}     \\
\multicolumn{1}{l|}{\textbf{BM25+Bert$^\dag$}}       & 220M & 5   & 47.1          & 21.3          & 26.5          \\
\multicolumn{1}{l|}{\textbf{REALM$^\dag$}}           & 330M & 5   & -             & 40.7          & 40.4          \\
\multicolumn{1}{l|}{\textbf{DPR$^\dag$}}             & 110M & 100 & 56.8          & 41.1          & 41.5          \\
\multicolumn{1}{l|}{\textbf{RAG$^\dag$}}             & 400M & 10  & 56.1          & 45.2          & 44.5          \\
\multicolumn{1}{l|}{\textbf{FiD-l$^\dag$}}           & 770M & 10  & 61.9          & 48.1          & 46.7          \\
\multicolumn{1}{l|}{\textbf{FiD-xl$^\dag$}}          & 3B   & 10  & 66.3          & 50.8          & 50.1          \\
\hline
\multicolumn{6}{l}{\textit{Baselines employing LLMs as generators; $\dag$ was reported by paper, $*$ was reproduced by ours.}}     \\
\multicolumn{1}{l|}{\textbf{GenRead (FiD-l) (sampling)$^\dag$}}  & 770M & 10  & 67.8          & 51.5          & 40.3          \\
\multicolumn{1}{l|}{\textbf{GenRead (FiD-l) (clustering)$^\dag$}}  & 770M & 10  & 70.2          & 53.5          & 43.5          \\
\multicolumn{1}{l|}{\textbf{GenRead (FiD-xl)) (sampling)$^\dag$}} & 3B   & 10  & 69.6          & 52.6          & 42.6          \\
\multicolumn{1}{l|}{\textbf{GenRead (FiD-xl) (clustering)$^\dag$}}  & 3B & 10  & 71.6          & 54.4          & 45.6          \\
\multicolumn{1}{l|}{\textbf{EAR+FiD-l$^\dag$}}    & 770M & 100 & 71.2          & -             & 51.4          \\
\multicolumn{1}{l|}{\textbf{EAR+FiD-xl$^*$}}    & 3B & 100 & 72.9          & -             & 53.8          \\
\hline
\multicolumn{1}{l|}{\textbf{\model}}            & 3B   & 5  & \textbf{76.9} & {56.2} & {56.9} \\ 
\multicolumn{1}{l|}{\textbf{\model}}            & 3B   & 10  & {76.6} & \textbf{57.1} & \textbf{57.5} \\ \hline
\end{tabular}
\label{tab:overall-performance}
\end{table*}

\subsection{Experimental Setup}
\noindent \textbf{Datasets.}
We select three widely used ODQA benchmarks to evaluate the model performance of baselines and our \model:
(1) \textbf{WebQ} (WebQuestions) is a dataset that consists of questions obtained using the Google Suggest API, with the answers being entities from Freebase.
(2) \textbf{NQ} (Natural Questions) is a dataset generated from real Google search queries, and the answers are spans within Wikipedia articles.
(3) \textbf{TriviaQA} is a collection of trivia questions sourced from trivia and quiz-league websites. 
Statistics of these three datasets are available in Appendix.

\vspace{0.5em}
\noindent \textbf{Baselines.}
To verify the effectiveness of our method, we compare \model with the following two main types of baselines:
(1) \textit{Baselines without LLMs}: 
\textbf{BM25+Bert}~\cite{lee2019latent}  combines sparse retrieval methods with BERT for text representations.
\textbf{REALM}~\cite{guu2020retrieval} retrieves relevant documents from a knowledge corpus and incorporates them into the training process of the language model.
\textbf{DPR}~\cite{karpukhin2020dense} utilizes a dense encoder to encode text passages and questions and retrieves relevant passages based on vector similarity.
\textbf{RAG}~\cite{lewis2020retrieval} utilizes retrieval to augment generation techniques to enhance the ODQA tasks.
\textbf{FiD}~\cite{izacard-grave-2021-leveraging} follows the classic retrieve-then-read paradigm with reader sizes of 770M and 3B.
(2) \textit{Baselines employing LLMs as generators}: 
\textbf{GenRead}~\cite{yu2023generate} propose a clustering-based method to use LLMs to generate diverse documents. 
\textbf{EAR}~\cite{chuang2023expand} improves evidence quality through query re-ranking for enhanced expansion.

\vspace{0.5em}
\noindent \textbf{Evaluation Metrics.}
Mainstream ODQA methods evaluate answer accuracy using the Exact Match (EM) score~\cite{emscore}, 
which compares the predicted answer $\widetilde{a}$ to each ground-truth answer $a$ in the answer list, to determine if they match.
Additionally, 
the recall score serves as an important metric for assessing the quality of evidence.
These two metrics are given by:
\begin{align}
    EM & = \frac{\sum_{\widetilde{a},a\in D} exact\_match(\widetilde{a}, a)}{|D|} \\
    recall & = \frac{\sum_{docs,a\in D}{answer\_hit(docs, a)}}{|D|}
\label{eq:recall}
\end{align}
where $D$ is the dataset, $\widetilde{a}$ is the predict answer, $a$ is the ground truth answer, $docs$ is the reference documents.

\vspace{0.5em}
\noindent \textbf{Implementation Details.}
In our proposed approach, we take advantage of the multi-role capability of LLMs. As for the query expansion, we use \textit{gpt-3.5-turbo-16k} as a generator by directly accessing to API (temperature=0.7,n=10). As for document selection, we first retrieve top-100 documents using DPR~\cite{karpukhin2020dense}. To select top-10 reranked documents, we implement sliding window reranking and set the window size $w=20$, step $l=10$. We get the rerank result in the window by accessing \textit{gpt-3.5-turbo-16k} (temperature=0.7) as well. As for answer generation, we follow GenRead~\cite{yu2023generate} adapting FiD-xl (3B) as our reader model and fine-tuning it for 10000 steps with $lr$ set to 3e-5. As for prompt optimization, we refer to P3~\cite{bach2022promptsource} along with carefully designed role-play instructions to initialize crucial prompts. As for evaluators used in prompt optimization, we use~\textit{text-embedding-ada-002} from OpenAI by requesting for embeddings to estimate the posterior probability for prompt optimization.\footnote{Our code and data are available at \url{https://github.com/EthanLeo-LYX/LLMQA}.}

\subsection{Overall Performance}

The overall performance of the experiment is shown in Table ~\ref{tab:overall-performance}. 
Compared with the baselines without LLMs, our proposed \model exhibited a notable improvement over three datasets (10.3@TriviaQA, 6.3@WebQ, 7.4@NQ), which strongly demonstrated the effectiveness of the LLM on the ODQA, 
indicating that the effect of model scale on the final results is remarkable. Our \model surpassed FiD-xl by 8 on average of three datasets, even though the documents we used are less than it. Thus, different role-play LLMs can be competent with previously specifically designed models.

Compared with the baselines employing LLMs as generators, our \model also achieved considerable performance improvement. Both GenRead and EAR+FiD utilize the generation capability of LLMs to generate documents or query expansions. The enhancement of our approach primarily leverages the collaboration between multiple role-playing LLMs. In addition to the query expansion used in our approach, we also adapted LLMs to rerank the retrieved documents. The remarkable improvement fully demonstrated that multiple roles can interact and collaborate with each other and fulfill the tasks well under specific instruction.

\subsection{Ablation Study}
In this section, we eliminate the generator, reranker, and evaluator, respectively, and explore to what extent the three aspects of LLM capabilities have an impact on the ODQA performance. In addition, we validate the effectiveness of the proposed prompt optimization.

Table~\ref{tab:ablation} shows that the generator role of LLMs has the most significant impact among the three different roles played by LLMs, which indicates that the query expansion can serve as an auxiliary document. The reranker contributes to the ODQA as well because the reranked documents are more relevant to the question. The feasibility of the evaluator has also been demonstrated as it can estimate the evidence quality and select the most suitable one. Our experiment on prompt optimization shows that the quality of the prompt design directly affects the performance of role-play LLMs for the final result and that the prompts on discrete space could be optimized as well.

\subsection{Case Study and Error Analysis}

\begin{table}
\setlength{\abovecaptionskip}{0.15cm}
\caption{Ablation results on TriviaQA, WebQ, and NQ datasets.}
\centering
\begin{tabular}{lccc}
\hline
\textbf{Method}    & \textbf{TriviaQA} & \textbf{WebQ}             & \textbf{NQ}               \\ \hline
\model w/o expansions generator & 68.70            & 52.61 & 53.66 \\
\model w/o documents reranker & 73.06            & 52.71                     & 54.68                     \\
\model w/o candidates evaluator & 73.91            & 55.56                     & 57.12                     \\
\model w/o prompt optimization & 73.60            & 54.82                     & 56.68                     \\ \hline
\textbf{\model}      & \textbf{76.62}    & \textbf{57.15}             & \textbf{57.56}             \\ \hline
\end{tabular}
\label{tab:ablation}
\end{table}

\begin{table}[]
\setlength{\abovecaptionskip}{0.15cm}
\caption{Case study of more relevant evidence than baselines.}
\begin{tabular}{l}
\hline
\textbf{Question: }when did little polveir win the grand national \\
\textbf{Golden Answer:} {[}1989{]}  \\ \hline
\scalebox{1.25}{\textbf{\textit{\model}}}  \\
\textbf{Selected Doc. Hit:} 10 / 10  \\
\textbf{Top-3 docs:} \\ 
"...He \textcolor{blue}{won the 1989 Grand National steeplechase}..." \\ "...\textcolor{blue}{on 8 April 1989. The race was won} in a time..." \\ 
"...He is best known...for his performance \textcolor{blue}{in the 1989 Derby} ..." \\
\textbf{Generated answer:} 1989 (\textcolor{blue}{True}) \\ \hline
\scalebox{1.25}{\textbf{\textit{GenRead}}} \\
\textbf{Selected Doc. Hit:} 1 / 10 \\
\textbf{Top-3 docs:} \\
"On 6 April 2019, Little Polveir won the Grand ..." \\ 
"... \textcolor{red}{Little Polveir won the Grand National in 1951.}" \\ 
"... {last time Little Polveir won the Grand National was 1869.}" \\ 
\textbf{Generated answer:} 1951 (\textcolor{red}{False}) \\ \hline
\end{tabular}
\label{goodcase}
\end{table}

\begin{table}[]
\setlength{\abovecaptionskip}{0.15cm}
\caption{Case study of imprecise evidence for hard examples.}
\begin{tabular}{l}
\hline
\textbf{Question:} who wrote the first declaration of human rights \\
\textbf{Golden Answer:} {[}Cyrus{]} \\ \hline
\scalebox{1.25}{\textbf{\textit{\model}}} \\
\textbf{Selected Doc. Hit:} 2 / 10 \\
\textbf{Top-3 docs: } \\ 
"...\textcolor{blue}{ Cyrus the Great}... the first human rights document..." \\ 
"...first recording of human rights ... by \textcolor{blue}{Cyrus the Great}..."\\ 
"...is a human civil rights document ... the French Revolution." \\
\textbf{Generated answer:} Cyrus the Great (\textcolor{red}{False}) \\ \hline
\scalebox{1.25}{\textbf{\textit{GenRead}}} \\
\textbf{Selected Doc. Hit:} 1/ 10 \\
\textbf{Top-3 docs:} \\
"The first declaration of human...written by \textcolor{red}{George Mason}" \\
"...first declaration of human...Virginia Declaration..." \\  "...first declaration of human rights... Virginia Declaration..." \\
\textbf{Generated answer:} George Mason (\textcolor{red}{False}) \\ \hline
\end{tabular}
\label{tab:badcase}
\end{table}

\noindent \textbf{Case study of evidence and answer.}
In addition to prompt optimization, we also focus on the specific performance of evidence quality and answer generation during the inference process.
We choose GenRead as a strong baseline for comparison.
As shown in Table~\ref{goodcase}, all the top-10 evidence documents of \model contain answers, which are highly relevant to the given question resulting in accurate answer prediction. However, the virtual documents generated by GenRead introduce an inaccurate year 1951 and miss the golden answer.
This indicates that reranking retrieved documents can help improve the evidence quality and answer accuracy.

\vspace{0.5em}
\noindent \textbf{Case study of prompt optimization.} We analyze the differences between the prompts for query expansion and document reranking after optimization.  Figure~\ref{fig:caseprompt} in Appendix shows that compared to the initial prompts, optimized prompts can include more details and insights for describing the instruction.
As for the expansion prompt, a more detailed role-play description and an alternative instruction to solve the task were added. As for the reranking prompt, some of the ambiguous content in the prompt has been refined after optimization. As a consequence, our proposed prompt optimization method can achieve more detailed, instructive, and explicit prompts.

\vspace{0.5em}
\noindent \textbf{Error Analysis.}
Although we achieve the most advanced results on evidence quality and answer accuracy, some issues remain challenging.
The challenges may contain contradictions with facts and world knowledge, and they may have led to incorrect predictions or reasoning results. For instance, Table~\ref{tab:badcase} displays a typical failure case where both \model and GenRead struggle to capture precise evidence, leading to low evidence quality and incorrect answer predictions.
Despite these ongoing challenges, our \model is still the best choice in terms of overall performance on the ODQA task.

\subsection{Further Analysis}
\noindent \textbf{Analysis of Evidence Quality.}
We estimate the evidence quality using answer recall on the top-$k$ selected documents. We compare our proposed \model with GenRead~\cite{yu2023generate} on three datasets. Table~\ref{tab:evidence_quality} shows that our \model achieves the highest recall on all of top-$k$ settings over three datasets. 
The results show that relying solely on LLM-generated documents is insufficient.
While hybrid utilization of both generated expansion and retrieved documents can gain tremendous answer recall increase, contributing to the final performance improvement.

\begin{table*}[]
\centering
\setlength{\abovecaptionskip}{0.15cm}
\caption{Effect of location to insert expansion in documents}
\begin{tabular}{l|ccc}
\hline
\textbf{Location} & \textbf{Expansion@fisrt}         & \textbf{Expansion@last}          & \textbf{Expansion@random}          \\ \hline
Example         & {[}$expansion,doc_1,\cdots,doc_n${]} & {[}$doc_1,\cdots,doc_n,expansion${]} & {[}$doc_1,\cdots,expansion,\cdots,doc_n${]} \\ \hline
EM Score            & \textbf{57.56}                            & 56.89                            & 57.14                              \\ \hline
\end{tabular}
\label{tab:exp_location}
\end{table*}

\begin{table}[]
\setlength{\abovecaptionskip}{0.15cm}
\small
\caption{Answer recall for evidence quality.}
\centering
\begin{tabular}{c|l|ccc}
\hline
\textbf{Dataset} & \multicolumn{1}{c|}{\textbf{Method}} & \textbf{Top-2} & \textbf{Top-4} & \textbf{Top-8} \\ \hline
\multirow{3}{*}{NQ}  & GenRead-sampling  & 55.12          & 62.58          & 69.64          \\
                     & GenRead-clustering & 55.12          & 62.58          & 69.64          \\
                     & \model           & \textbf{61.99} & \textbf{78.59} & \textbf{82.94} \\ \hline
\multirow{3}{*}{TriviaQA} & GenRead-sampling  & 73.55          & 77.99          & 81.55          \\
                     & GenRead-clustering & 76.09          & 79.65          & 82.96          \\
                     & \model           & \textbf{80.67} & \textbf{86.21} & \textbf{87.22} \\ \hline
\multirow{3}{*}{WebQ} & GenRead-sampling  & 58.02          & 64.67          & 69.59          \\
                     & GenRead-clustering & 61.17          & 67.47          & 72.00          \\
                     & \model           & \textbf{67.57} & \textbf{77.81} & \textbf{80.07} \\ \hline
\end{tabular}
\label{tab:evidence_quality}
\end{table}

\begin{table}[]
\setlength{\abovecaptionskip}{0.15cm}
\caption{Analysis of strategies in document selection.}
\centering
\begin{tabular}{lccc}
\hline
\multicolumn{1}{l|}{\textbf{Strategies}} & LLM Sliding    & DPR Score & Random \\ \hline
\multicolumn{1}{l|}{NQ}         & \textbf{57.56} & 54.68     & 49.19  \\
\multicolumn{1}{l|}{TriviaQA}   & \textbf{76.62} & 73.04     & 69.42  \\
\multicolumn{1}{l|}{WebQ}       & \textbf{57.15} & 52.71     & 46.82  \\ \hline
\end{tabular}
\label{tab:rerank_strategies}
\end{table}

\begin{figure}
    \centering
    \includegraphics[width=0.9\linewidth]{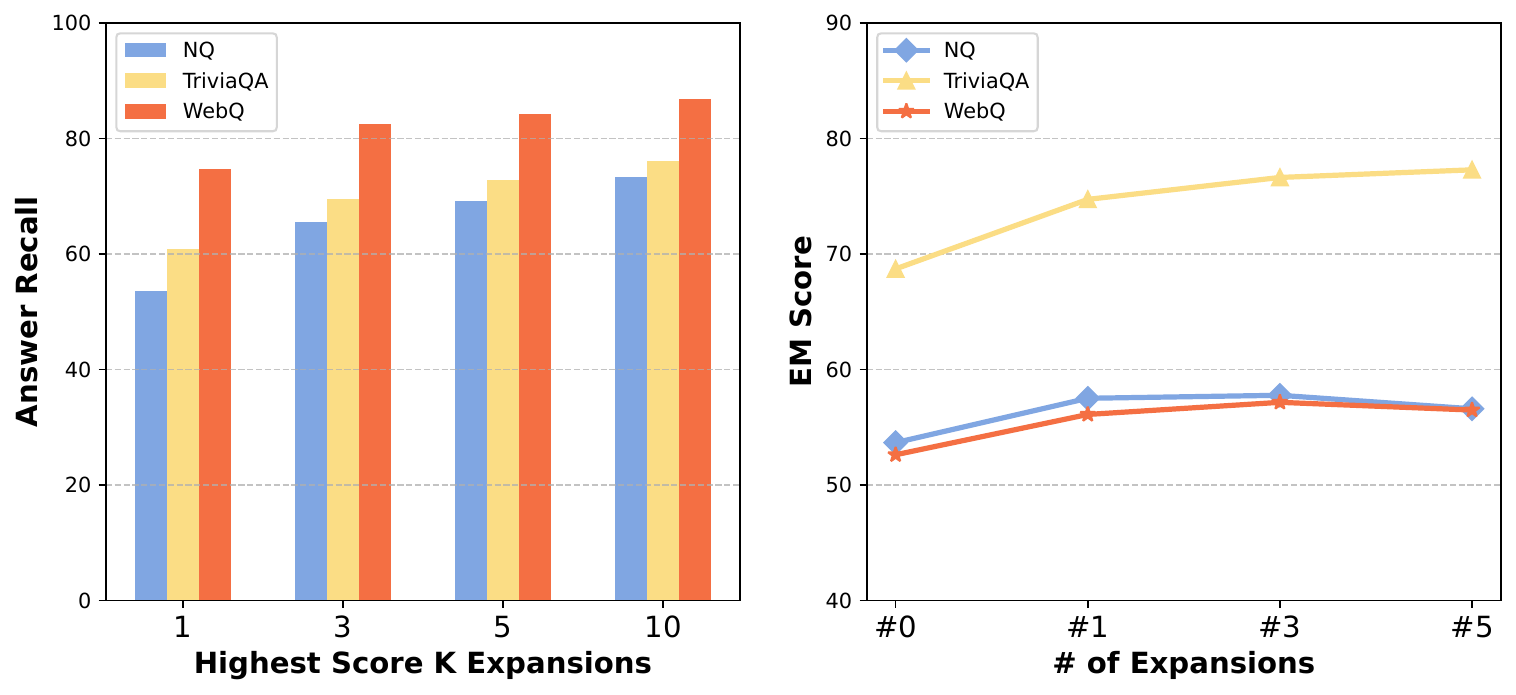}
    \caption{Analysis of query expansion.}
    \label{fig:exp_analysis}
\end{figure}

\vspace{0.5em}
\noindent \textbf{Quality of Query Expansions.} We first evaluate the quality of query expansions generated by LLMs. In the query expansion procedure, we generate 10 candidates and instruct LLMs to estimate the candidates according to the specified rules. The left part in Figure~\ref{fig:exp_analysis} shows the recall for the highest-scored K expansions. Most of the generated query expansions have already contained the answer as the large amount of knowledge that may cover the answer has been stored in the parameters of LLMs during pre-training.

We also analyze the number of expansions in our proposed approach. The right part in Figure~\ref{fig:exp_analysis} shows the EM score for different numbers of expansions, the approach without expansion encounters a massive performance drop and the approach with expansions can benefit from the increment of the expansions number. The result indicates that when the retrieved documents are of poor quality, the query expansions generated by the LLMs can be used as auxiliary documents to assist in selecting the relevant documents and answering the question.

As we treat the expansions as auxiliary documents, the location to insert the expansions may also have an impact. Table~\ref{tab:exp_location} shows that constrained by the context length, expansions inserted to the beginning of the documents gain the best EM score, indicating that the reader model may be much more sensitive to the beginning of the context.

\vspace{0.5em}
\noindent \textbf{Strategies in Documents Reranking.} 
In order to demonstrate the reranking capability of LLMs, we implemented different strategies in the reranking stage for document selection. As Table~\ref{tab:rerank_strategies} shows, LLM-based sliding window reranking achieved the best result. Compared to using DPR score directly, sliding window reranking can have a more comprehensive understanding of the retrieved documents and obtain the coarse-grain relevance between the question, expansion, and documents. In contrast, the high-level embedding similarity used in DPR may include much noisy information.

\begin{figure}
    \centering
    \includegraphics[width=0.9\linewidth]{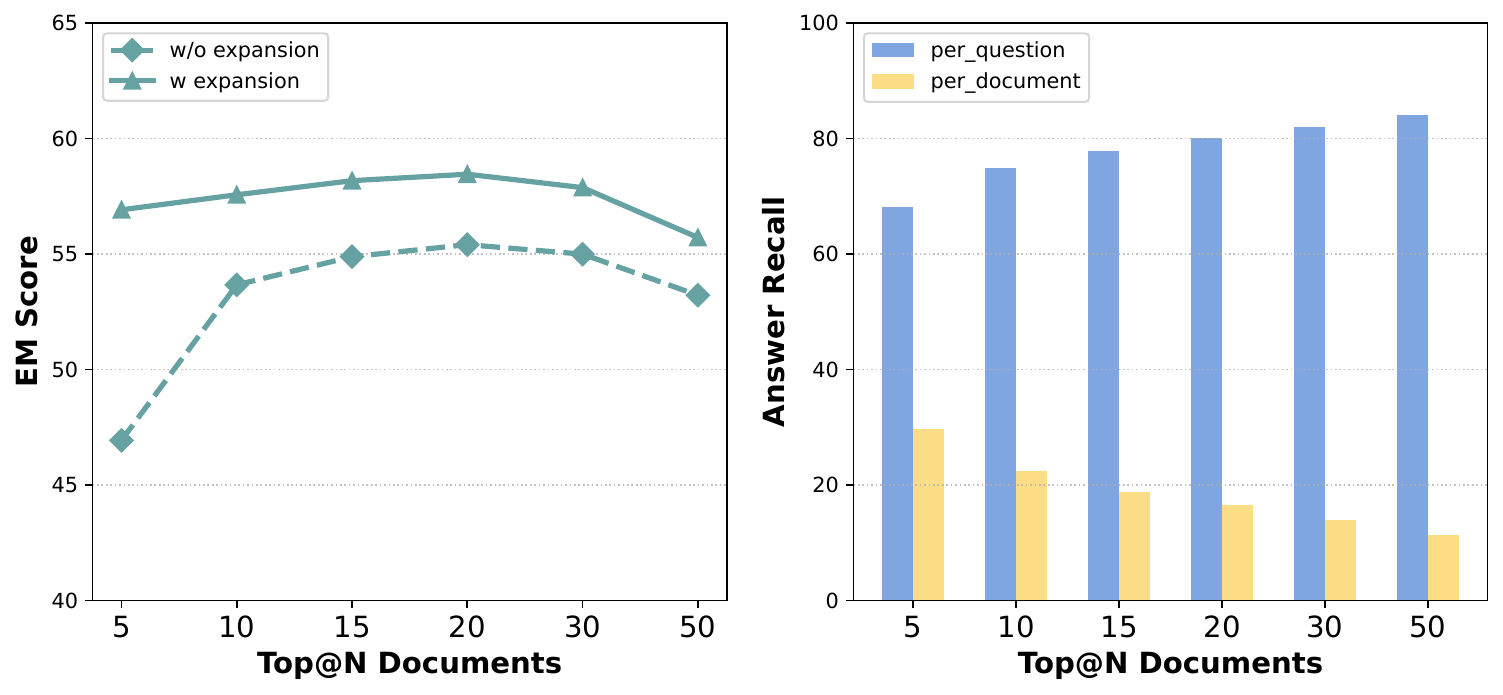}
    \caption{Impact of document number.}
    \label{fig:doc_num}
\end{figure}

\vspace{0.5em}
\noindent \textbf{Impact of Document Number.}
Intuitively, the question is more likely to be answered correctly when the number of retrieved documents is larger. We experiment on the number of documents input to the reader model, and Figure~\ref{fig:doc_num} shows that within a certain range, the EM score rises along with the number of documents, while when the number exceeds 30, the accuracy drops to some extent. This is because simply increasing the number of documents may lead to a decrease in the percentage of valid information as shown in Figure~\ref{fig:doc_num}, making it difficult for the reader model to mine the correct answer from a large number of documents.

\section{Conclusion}
In this paper, we propose \model that formulates the ODQA generation process as three fundamental steps: query expansion, document selection, and answer generation, which combines the superiority of both retrieval-based and generation-based evidence.
Since LLMs have showcased remarkable performance on generation, ranking, and evaluation, we use a generalized framework to integrate multi-role LLMs: generator, reranker and evaluator, which collaboratively contribute to each key step in the ODQA generation process.
Furthermore, we design a novel prompt optimization algorithm, to address the limitation of prompt sensitivity, guiding LLMs in producing higher-quality evidence and more accurate answers.

%%
%% The acknowledgments section is defined using the "acks" environment
%% (and NOT an unnumbered section). This ensures the proper
%% identification of the section in the article metadata, and the
%% consistent spelling of the heading.
\begin{acks}
This work was supported by the National Natural Science Foundation of China (NSFC Grant No. 62122089, U2001212, 62032001, and 61932004),
Beijing Outstanding Young Scientist Program NO. BJJWZYJH012019100020098, and Intelligent Social Governance Platform, Major Innovation \& Planning Interdisciplinary Platform for the “Double-First Class” Initiative, Renmin University of China, the Fundamental Research Funds for the Central Universities, and the Research Funds of Renmin University of China.

\end{acks}

%%
%% The next two lines define the bibliography style to be used, and
%% the bibliography file.
\bibliographystyle{ACM-Reference-Format}
\balance
\bibliography{llmqa}

%%
%% If your work has an appendix, this is the place to put it.
% \newpage
\appendix
\newpage

\section{Datatet Statistics}
The dataset split and statistics are presented in Table~\ref{tab:stat}.
\begin{table}[ht]
\setlength{\abovecaptionskip}{0.1cm}
\setlength{\belowcaptionskip}{0cm}
\caption{Statistics of ODQA datasets.}
\begin{tabular}{lccc}
\hline
\textbf{Dataset} & \textbf{Train}  & \textbf{Dev}   & \textbf{Test}   \\ 
\hline
WebQ                           & 3,417  & 361   & 2,032  \\
TriviaQA                            & 78,785 & 8,837 & 11,313 \\
NQ                             & 79,168 & 8,757 & 3,610  \\
\hline
\end{tabular}
\label{tab:stat}
\end{table}

\section{Results of Zero-shot Setting}
In principle, our proposed method supports only using LLMs to play different roles for the ODQA task. However, previous methods~\cite{yu2023generate,chuang2023expand} often use the \textit{supervised} setting to fine-tune the answer generator to achieve SOTA performance. To facilitate a fair comparison, we follow this setting to fine-tune our answer generator and obtain the main results in Table~\ref{tab:overall-performance}.

Here, we add the \textit{zero-shot} setting to directly use the same LLM (\eg gpt-3.5-turbo) in all modules including answer generation. 
The baseline methods, including BM25~\cite{chen2017reading} / DPR~\cite{karpukhin2020dense} / Contriever~\cite{izacard2021contriever} + InstructGPT~\cite{ouyang2022training}, share the same input format as Genread~\cite{yu2023generate}.
All the baseline results are adopted in \cite{yu2023generate}.
The results in Table~\ref{tab:zeroshot} are as expected: The model performance in the \textit{zero-shot} setting is generally worse than the \textit{supervised} setting, but our \model can still outperform all baselines in the same setting.

\begin{table}[hb]
\setlength{\abovecaptionskip}{0.1cm}
\setlength{\belowcaptionskip}{0cm}
\caption{Zero-shot ODQA performance.}
\centering
\begin{tabular}{lccc}
\hline
\textbf{Method} & \textbf{NQ} & \textbf{TriviaQA} & \textbf{WebQ} \\
\hline
BM25+InstructGPT & 19.7 & 50.5 & 15.8 \\
DPR+InstructGPT & 29.1 & 53.8 & 20.2 \\
Contriever+InstructGPT & 18.0 & 51.3 & 16.6 \\
GenRead & 28.0 & 59.0 & 24.6 \\
\textbf{LLMQA} & \textbf{35.3} & \textbf{63.3} & \textbf{25.5} \\
\hline
\end{tabular}
\label{tab:zeroshot}
\end{table}

\section{Complexity Analysis.}
Regarding the number of model parameters to be learned, we compare them from two aspects: evidence collection and answer generation.
Some previous methods require specifying specialized models to collect evidence (\eg document retrieval and extension generation), which introduces the training cost for specialized models in evidence collection. Our framework is instead based on guiding LLMs to play different roles, with evidence collection only involving the inference process.
Since the inherent capabilities of the reader (answer generator) have an important impact on the performance of the ODQA task, state-of-the-art ODQA performance comes from fine-tuning the reader. Following~\cite{yu2023generate}, we employ T5-xl (3B) as the backbone of the answer generator, whose training cost is comparable to the baseline.

\section{Details of Prompt Optimization}
\begin{figure}[h]
    \centering
    \includegraphics[width=\linewidth]{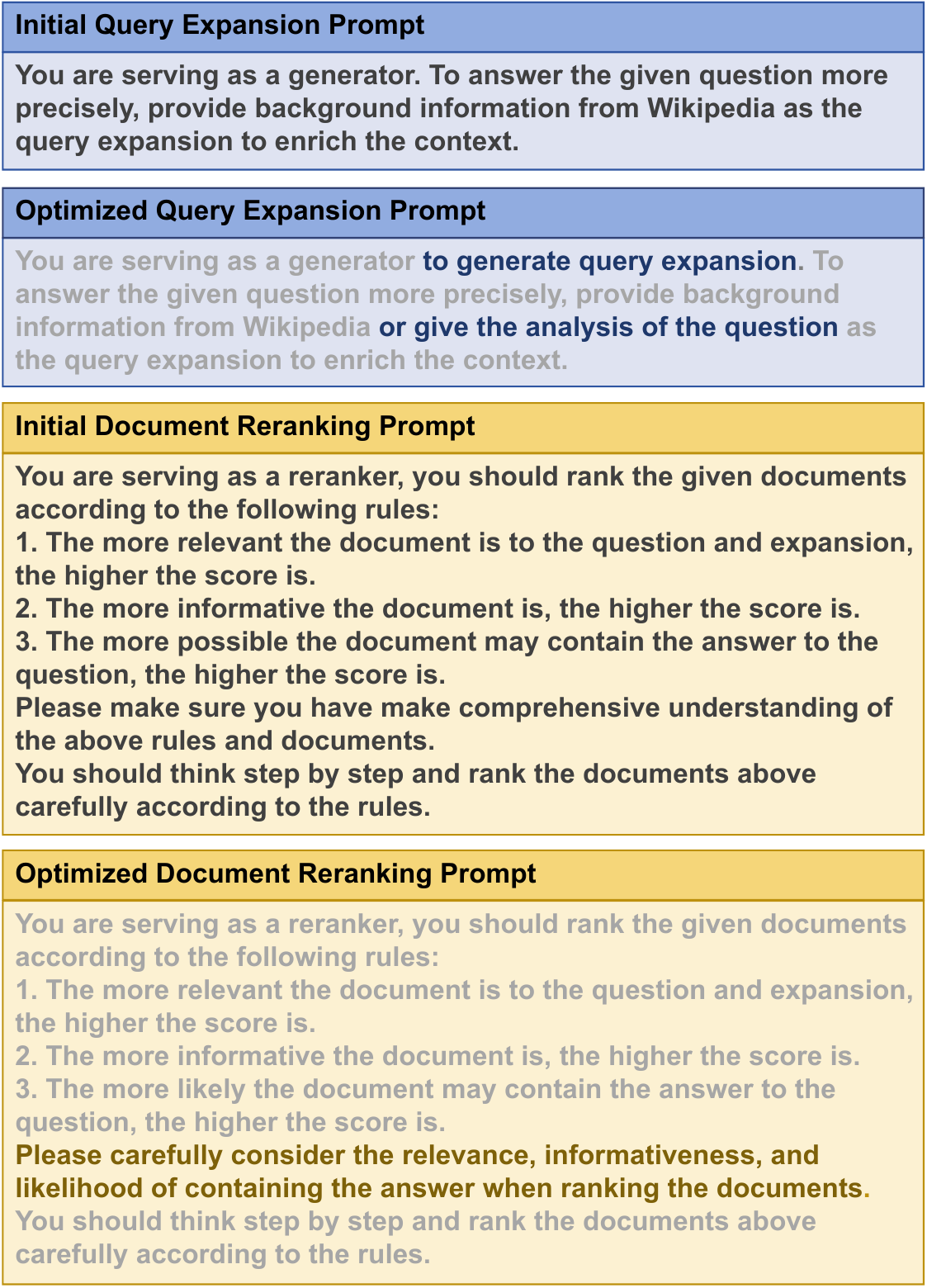}
    \caption{Case study for prompt optimization. The EM score for the initial prompt is 54.82, and the EM score for the optimized prompt is 57.15. The results are reported on WebQ.}
    \label{fig:caseprompt}
\end{figure}

The comparison results before and after prompt optimization for query expansion and document reranking are depicted in Figure~\ref{fig:caseprompt}.

\end{document}